\pdfoutput=1

\documentclass[11pt]{article}

\usepackage[colorlinks=true,citecolor=blue]{hyperref}

\usepackage{times}
\usepackage{latexsym}
\usepackage{booktabs}       %
\usepackage{multirow}
\usepackage{graphicx}
\usepackage{color}
\usepackage{caption}
\usepackage{subcaption}
\usepackage{algorithm}
\usepackage{algpseudocode}
\newif\ifarxiv
\arxivtrue

\ifarxiv
         \usepackage{lineno}
        \usepackage{url}
        \usepackage{natbib}
        \usepackage{color}
        \usepackage{xcolor}
\else
     \usepackage[final]{acl}
\fi

\usepackage[T1]{fontenc}

\usepackage[utf8]{inputenc}

\usepackage{microtype}
\usepackage{amsmath}
\usepackage{amsthm}
\usepackage{amssymb}
\usepackage{geometry}
\usepackage{enumitem}

\usepackage{inconsolata}
\usepackage{dsfont}
\theoremstyle{definition}

\author{%
  {Michal Lukasik ~~~~~~~~Harikrishna Narasimhan}\\
{{Aditya Krishna Menon ~~~~~~~~Felix Yu ~~~~~~~~Sanjiv Kumar}} \\
  \texttt{\{mlukasik, hnarasimhan, adityakmenon, felixyu, sanjivk\}@google.com} \\
  Google Research, USA
}

\usepackage{amsmath}
\DeclareMathOperator*{\argmax}{argmax}

\title{Regression-aware Inference with LLMs}
\begin{document}
\maketitle
\begin{abstract}
Large language models (LLMs) have shown strong results on a range of applications, including regression and scoring tasks.
Typically, one obtains outputs from an LLM via autoregressive sampling from the model's output distribution. 
We show that this inference strategy can be sub-optimal for common regression and scoring 
evaluation metrics. 
As a remedy, we build on prior work on Minimum Bayes Risk decoding,
and propose alternate inference strategies that estimate the  Bayes-optimal solution for regression and scoring metrics in closed-form  from sampled responses.
We show that our proposal significantly improves over baselines across datasets and models.
\end{abstract}

\section{Introduction}
\label{s:introduction}
Large language models (LLMs) are currently the most capable models across many NLP tasks~\citep{openai2023gpt4,palm2,touvron2023llama,geminiteam2023gemini}.
Owing to their remarkable \emph{few-} and \emph{zero-shot} abilities \citep{Wei:2022,kojima2023large}, pre-trained LLMs 
are often applied 
without \emph{any} additional %
training on 
domain-specific
datasets:
instead,
one may query the %
LLM
with a suitably crafted input prompt.

More recently, LLMs have been successfuly applied to regression and scoring tasks.
For example, \citet{gruver2023large} explored zero-shot learning for time series prediction; \citet{vacareanu2024words} showed how LLMs are remarkably strong at in-context learning for %
regression tasks; \citet{liu2023goat,yang2023gpt} considered autoregressive fine-tuning over numerical targets %
applied to arithmetic tasks; and \citet{qin2023large} applied LLMs for listwise ranking.

The quality of an LLM is often assessed using an application-specific \emph{evaluation metric}.
One popular metric is the
\emph{exact match} (EM), which penalises \emph{any} response not exactly equal to the one in the dataset annotation. 
This is an analogue of the conventional {classification accuracy}.
While EM is an intuitive metric, 
there are many applications where it is not suitable.
This 
includes
tasks such relevance scoring \citep{Cer_2017} and sentiment analysis \citep{NIPS2017_c86a7ee3}, where  the outputs are numerical or ordinal categories. 
 In these cases, one instead prefers
metrics such as the squared error, absolute error or ranking scores that take the outputs' ordinal nature into account. %

Despite the wide variety of evaluation metrics, LLM \emph{inference} is typically performed in the same manner for \emph{every} task:
namely, one performs auto-regressive sampling from the LLM's underlying distribution (see \S\ref{s:problem}).
While intuitive, such inference  does not explicitly consider the downstream evaluation metric of interest.
This raises a natural question: \emph{is there value in adapting the inference procedure to the evaluation metric at hand for regression and scoring tasks}?

A prominent line of work takes a decision-theoretic approach to the above problem. Dubbed as \emph{Minimum Bayes Risk} (MBR) decoding, this approach seeks to optimize at inference time the metric of choice under the model's distribution \citep{bickel1977mathematical,kumar-byrne-2004-minimum,eikema-aziz-2020-map,bertsch-etal-2023-mbr}. 
Much of the work on MBR is focused on evaluation metrics for machine translation and text generation tasks, such as the BLEU score \citep{papineni2002bleu}.  Of particular interest in this literature are self-consistency based decoding strategies that take a (weighted) majority vote of sampled responses \citep{wang2023selfconsistency}, which have shown to provide quality gains in arithmetic and reasoning problems. %

\begin{figure}[t]
\vspace{-2pt}
\centering
     \resizebox{0.8\linewidth}{!}{
        \includegraphics[scale=0.7]{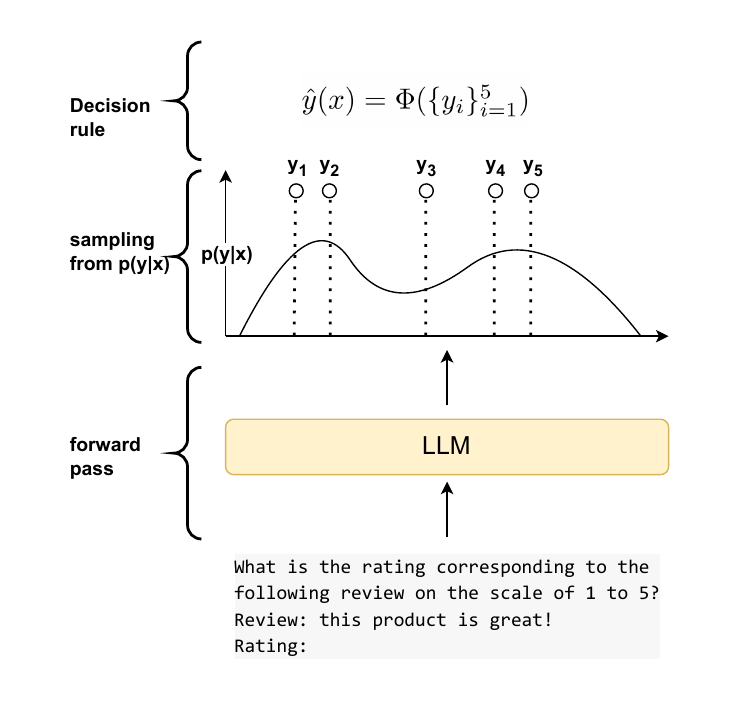}
        }
     \caption{Illustration of metric-aware LLM inference %
     for regression and scoring tasks. %
     An input $x$ is passed to the LLM, and samples are drawn from the distribution over targets $y$ conditioned on $x$.  %
     These are then used to find the target optimizing a metric ${m}$ through a closed-form decision rule $\Phi$ (e.g., mean or median); %
      Table~\ref{tbl:decision_rules} presents specific solutions across metrics.
     }
    \label{fig:illustrative}
\end{figure}

In this paper, we %
build on the existing literature on MBR to design metric-aware inference strategies for \emph{general regression and scoring} tasks.
We first observe that choosing the most likely target for an  input corresponds to \emph{inherently optimizing for the EM} metric, and is consequently \emph{not optimal} when EM is not the metric of choice. 
As a remedy, we propose estimating the Bayes-optimal output for a metric under the model's distribution (see Figure~\ref{fig:illustrative} for an illustration of our method); we show that this admits a
\emph{closed-form} solution for common regression and ranking metrics, and only requires estimating a simple statistic from the sampled responses.
In contrast, prior MBR methods for translation and summarization often require heuristically solving an intractable maximization problem \citep{ehling2007minimum, bertsch-etal-2023-mbr}. 
We show across datasets and models how our approach yields gains over choosing the most likely target, and over self-consistency based approaches.

  \begin{algorithm*}
   \caption{RAIL: Regression-aware Inference with LLMs}
      \label{a:RAIL_ALGO}
    \begin{algorithmic}[1]
    \State \textbf{input:} Model $M$, \#samples $K$, sampling temperature $T$, effective temperature  $T'$, decision rule $\Phi$
        \For{$i = 1$ to $K$}
            \State $\hat{y}_i, \hat{p}_i = M.\text{\tt generate}(T)$\Comment{{\color{gray}$M$.{\tt generate} returns the sampled target and its probability.}}
        \EndFor
        \State $\alpha = \frac{T}{T'} - 1$\Comment{{\color{gray}Compute post-hoc temperature scaling so that the effective temperature used is $T'$}}
        \State \textbf{return} $\Phi(\hat{y}_1, \ldots, \hat{y}_K, \hat{p}_1, \ldots, \hat{p}_K, \alpha )$
\end{algorithmic}
\end{algorithm*}

\vspace{-5pt}
\section{When (na\"{i}ve) LLM inference fails on regression tasks}
\label{s:problem}
We begin with the problem setting. 
For a finite vocabulary $V$ of \emph{tokens} (e.g., words in English), let $D$ denote a distribution over \emph{inputs} $x \in X \subseteq {V}^*$ comprising of strings of tokens, 
and \emph{targets}  $y \in Y$. Let $p(y \,|\, x)$ denote the conditional distribution over targets given an input. 
We consider a special case of this setting where $Y \subset \mathbb{R}$ corresponds to numeric targets.
Here, we assume that each $y \in Y$ has a unique string representation ${\tt str}( y ) \in V^*$; for example, the integer $1$ has the string encoding {\tt"1"}.

A \emph{language model} (LM) takes a string $x$ as input and predicts an output $\hat{y} \in V^*$. Typically, the LM first produces a distribution $\hat{p}( \cdot \,|\, x )$ over targets.
In a slight abuse of notation, we use $\hat{p}( y \,|\, x ) \stackrel{\cdot}{=} \hat{p}( {\tt str}( y ) \,|\, x )$ to denote the conditional probability of a numerical output $y$ given input $x$.
Note that even for problems where numerical targets are expected, an LM may return a non-zero probability to non-numerical targets.

A prediction from an LM is typically derived via a suitable \emph{inference} (or \emph{decoding})  procedure. 
Perhaps the most common inference strategy is to choose the mode of $\hat{p}( \cdot \,|\, x )$:
\begin{equation}
\label{eq:greedy-decoding}
    \hat{y}(x) := \argmax_{y \in V^*} \hat{p}(y \,|\, x).
\end{equation}
In practice, one may approximate the mode via greedy decoding or beam search, or sampling multiple candidates %
and picking the among them the one with the highest likelihood score~\citep{Naseh:2023}.
In principle, the extracted target may not be numerical.
In such cases, a possible strategy is to resort to predicting a default numerical value such as $0.0$.
In practice, we find the targets from high-quality LLMs tend to be numerical even under zero-shot settings, and so converting most likely targets from $V^*$ to $Y$ is usually possible.

The quality of an LM's prediction is measured by some \emph{evaluation metric} $m( y, \hat{y} )$, where we assume that \emph{higher} values are \emph{better}.
While the  \emph{exact match} (EM), given by $m( y, \hat{y} ) = \mathds{1}( {y = \hat{y}} )$, is a commonly used evaluation metric, there are a range of other metrics popularly used to evaluate LMs. These include the (negative) squared error $m( y, \hat{y} ) = -(y - \hat{y})^2$ or absolute error $m( y, \hat{y} ) = -|y - \hat{y}|$ for regression tasks.
A natural goal is to then choose the inference strategy $\hat{y}(x)$ to maximize the metric $m$ of interest, i.e., to maximize the expected utility:
\begin{align}
    \mathbb{E}_{(x, y) \sim D}\left[ m(y, \hat{y}(x)) \right].
    \label{eq:expected-utility}
\end{align}
For many choices of metric $m(y, \hat{y}(x))$, 
picking the mode of the predicted distribution \eqref{eq:greedy-decoding} can be  sub-optimal for \eqref{eq:expected-utility}.
As an example, consider predicting the star rating (on the scale $1$--$5$)
associated with a review text.
Suppose $m(y, \hat{y})$ is the negative absolute error between the true and predicted ratings.
Given the review text {\tt ``This keybord is suitable for fast typers''}, 
suppose the LM responses and the associated probabilities are \{{\tt``1''}: 0.3,\, {\tt``2''}: 0.0,\, {\tt``3''}: 0.3,\, {\tt``4''}: 0.0,\, {\tt``5''}: 0.4\}. 
The mode of the predicted probabilities is {\tt``5''}. %
In contrast, the maximizer of  \eqref{eq:expected-utility} is the median {\tt``3''}.

\begin{figure*}[!t]
\centering
    \begin{subfigure}[t]{0.3\textwidth}
        \centering
        \includegraphics[scale=0.165]{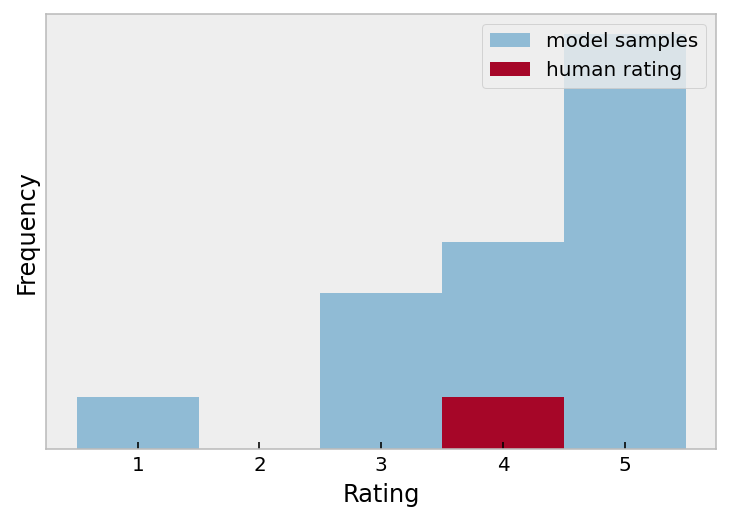} %
        \caption{
        \emph{It is a nice color of black and my husband likes how it feels in his hand.}}
    \end{subfigure}
    \hspace{10pt}
    \begin{subfigure}[t]{0.3\textwidth}
        \centering
        \includegraphics[scale=0.165]{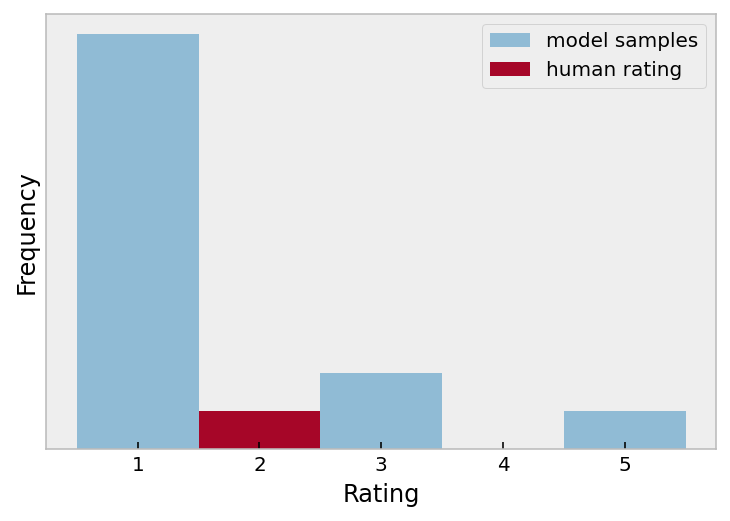} 
        \caption{
        \emph{This item is a good idea. However, Unless the ear canal is reasonably deep (...) it's of no use. The plastic hooks that come with it are hard and too small (...). Might be good for children.}}
    \end{subfigure}
    \hspace{10pt}
    \begin{subfigure}[t]{0.3\textwidth}
        \centering
        \includegraphics[scale=0.165]{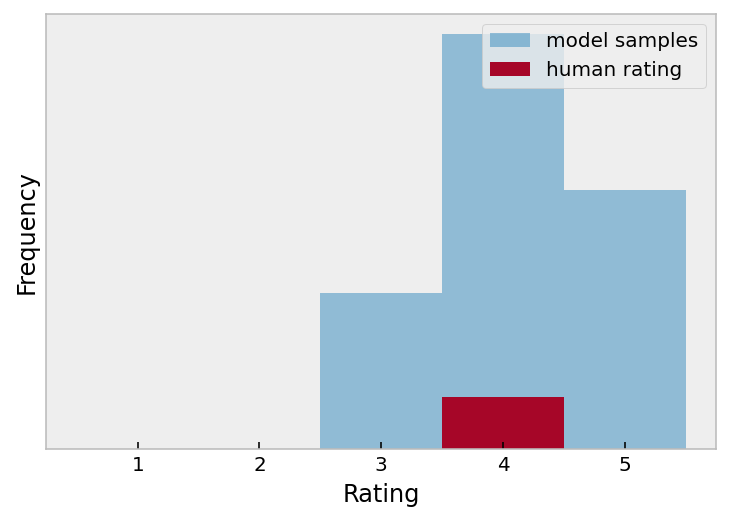} 
        \caption{
        \emph{One of the sides is made for apple products, the other is just standard usb. Both will work with apple products, just one side (the A side) charges faster. Other than that, it's fantastic. :D}}
    \end{subfigure}
     \caption{
     Examples from the Amazon dataset and the corresponding: human annotations and samples from the model. We find that in many cases, taking into account the model distribution (i.e. a \emph{mean} of the distribution) allows for a prediction closer to the annotation than simply taking the \emph{mode} of the distribution.
     }
    \label{fig:examples_amazon}
\end{figure*}

In Figure~\ref{fig:examples_amazon}, we report examples from the Amazon dataset and the corresponding human annotations and samples from the model. 
Notice how samples cover significant proportions of the ratings. 
We find that the samples end up in the vicinity of the human annotation, and thus in many cases taking a \emph{mean} over samples helps improve the prediction over the \emph{mode}.

\begin{table*}
    \centering
    \renewcommand{\arraystretch}{1.25}
\resizebox{\linewidth}{!}{
    \begin{tabular}{@{}llllll@{}}
        \toprule
         \textbf{Problem} & \textbf{Label} %
         & \textbf{Pred.}%
         & \textbf{Metric} & \textbf{Optimal rule} &
         $\Phi(\hat{y}_1, \ldots, \hat{y}_K, \hat{p}_1, \ldots, \hat{p}_K, \alpha )$
         \\ %
\toprule
Classification & $[K]$ &
$[K]$
&
$\mathds{1}( {y = \hat{y}} )$ & $\hat{y}(x) := \argmax_y\, p(y \,|\, x)$ & %
$\hat{y}_i$ s.t. $i=\argmax_{j}\, \hat{p}_{j}$
\\
Regression & $\mathbb{R}$ & $\mathbb{R}$ & $-(y - \hat{y})^2$ & $\hat{y}(x) := \mathbb{E}_{y \sim p(\cdot \,|\, x)} [y]$ & 
$\sum_{i} \frac{\hat{p}_i^{\alpha} \cdot \hat{y}_i}{\sum_{j}\hat{p}_j^{\alpha} \cdot \hat{y}_j}$
\\
Ordinal regression & $[K]$ & $[K]$ & $-|y - \hat{y}|$ & $\hat{y}(x) := \text{median}[ p(\cdot \,|\, x) ]$ & 
$\hat{y}_{i}$ s.t. $i=\text{median}( \hat{p}^\alpha_1, \ldots, \hat{p}^\alpha_K )$\\
Bi-partite ranking & $\pm 1$ & $\mathbb{R}$ &  $\text{AUC}$ ($c_{y, y'} = 1$) & 
 $\hat{y}(x) := p(y=+1|x)$ & 
$\sum_{i} \frac{\hat{p}_i^{\alpha} \cdot \mathds{1}(\hat{y}_i = 1) }{\sum_{j}\hat{p}_j^{\alpha} \cdot \mathds{1}(\hat{y}_j = 1)}$
\\
Multi-partite ranking & $[K]$ & $\mathbb{R}$ & 
{$\text{AUC}$ %
($c_{y, y'} = |y - y'|$)}
& $\hat{y}(x) := \mathbb{E}_{y \sim p(\cdot \,|\, x)} [y]$ & 
$\sum_{i} \frac{\hat{p}_i^{\alpha} \cdot \hat{y}_i}{\sum_{j}\hat{p}_j^{\alpha} \cdot \hat{y}_j}$
\\
  \bottomrule
  \\[-5pt]
\end{tabular}
}
    \caption{Optimal decision rule for varying: label space, model prediction space and evaluation metric. We denote $[K] = \{1, \ldots, K\}$. %
    The final column shows the empirical rule as a function of sampled outputs, corresponding scores, and a rescaling temperature $\alpha$ (see Section~\ref{ssec:post-hoc-temp}).
    }
    \label{tbl:decision_rules}
    \vspace{-10pt}
\end{table*}

\section{Metric-aware LLM inference}
\label{s:dec_theory}
\subsection{Minimum Bayes risk decoding}
We seek to design decoding strategies that maximize the expected utility in \eqref{eq:expected-utility}. 
Ideally, if we had access to the true conditional probabilities $p(\cdot \,|\, x)$, 
the maximizer of \eqref{eq:expected-utility} is given by:
\begin{align}
    \hat{y}^*(x) \in \argmax_{y' \in Y}\, \mathbb{E}_{y \sim p(\cdot \,|\, x)}\left[ m(y, y') \right].
    \label{eq:bayes-opt}
\end{align}
When $m$ is the EM metric, the optimal inference strategy is $\hat{y}^*(x) \in \argmax_{y \in Y}\, p(y \,|\, x),$ which is what 
common approaches such as greedy decoding seek to approximate. 

In general, however, the optimal decoding strategy can have a very different form, and the mode of $p(\cdot | x)$ has been shown to be suboptimal on generation tasks \citep{eikema-aziz-2020-map}. 
For example, as shown in Table \ref{tbl:decision_rules}, for evaluation metrics over numerical targets such as the squared error or the absolute error, the optimal inference strategy is to take the mean or median of $p(\cdot|x)$ \citep{bishop2006}. %

\vspace{-5pt}
\subsection{Closed-form optimal solution}
\label{s:approx_bayes_opt}
 
In practice, we mimic the Bayes-optimal solution in \eqref{eq:bayes-opt} 
with two approximations.
First,
we replace the true conditional distribution ${p}(\cdot \,|\, x )$ with the LM's predicted distribution $\hat{p}(\cdot \,|\, x )$. 
This is a reasonable approximation when the LM is pre-trained
with next-token prediction
objective based on the softmax cross-entropy loss;
the latter is a strictly proper loss, 
whose minimizer under an unrestricted hypothesis class is the true conditional distribution $p(y \,|\, x)$~\citep{Gneiting:2007}.
Second, we estimate the expectation in \eqref{eq:bayes-opt} by sampling  $K$ outputs from $\hat{p}(\cdot \,|\, x )$, %
and then computing:
\begin{align}
    \hat{y}(x) \in \argmax_{y' \in Y}\, \sum_{i=1}^K m(y_i, y').
    \label{eq:bayes-opt-empirical}
\end{align}
Even with these approximations, maximizing \eqref{eq:bayes-opt-empirical} over all outputs $Y$ is intractable in general.
Prior literature on MBR for metrics like BLEU heuristically perform maximization over a small set of candidates  \citep{ehling2007minimum, bertsch-etal-2023-mbr}. 

In this paper, we consider regression and scoring metrics, for which the above maximization can be computed in \emph{closed-form}.
As shown in Table \ref{tbl:decision_rules}, these solutions can be estimated by computing simple statistics  from the sampled responses, such as the sample mean $\hat{y}(x) = \frac{1}{K}\sum_{i=1}^K y_i$ for the squared error.
We refer to this approach as \textbf{R}egression-\textbf{a}ware \textbf{I}nference with \textbf{L}LMs (\textbf{RAIL}).

\subsection{Post-hoc temperature scaling}
\label{ssec:post-hoc-temp}
When sampling from $\hat{p}(\cdot \,|\, x)$, it  often helps to apply temperature scaling to the LM logits to control the sampled outputs' diversity. 
This is particularly important in our procedure, where we wish to approximate expectations over $\hat{p}(\cdot|x)$ using a few samples.

In practice, one may sample from $\hat{p}(\cdot \,|\, x)$ with temperature $T=1$, and apply temperature scaling in a post-hoc manner by employing a weighted version of the objective in \eqref{eq:bayes-opt-empirical}:
\begin{align}
    \hspace{-5pt}\hat{y}(x) \in \argmax_{y' \in Y}\, \sum_{i=1}^K \left(\hat{p}(y_i | x)\right)^\alpha \cdot m(y_i, y'),\hspace{-2pt}
    \label{eq:bayes-opt-empirical-weighteed}
\end{align}
where $\alpha$ can be seen as the temperature scaling parameter. The above summation is a (scaled) estimate of 
$\mathbb{E}_{y \sim \hat{p}(\cdot \,|\, x)} \left[\hat{p}(y \,|\, x)^\alpha \cdot m(y, y')\right]$. For probabilities  $\hat{p}(y_i \,|\, x) \propto \exp( f(x, y_i) )$ defined by logits $f(x, y_i)$, this is equivalent to computing the expectation under the temperature-scaled distribution $\hat{p}_{\alpha}(y \,|\, x) \propto \exp( (1 + \alpha) \cdot f(x, y) )$, \emph{modulo} a normalization factor. 
We consider an analogous weighting scheme for the plug-in estimators of the closed-form solutions  in Table~\ref{tbl:decision_rules}.

Algorithm   \ref{a:RAIL_ALGO}
outlines the RAIL procedure, with both a sampling temperature $T$ and an effective temperature $T'$ as inputs. The algorithm first draws samples from an LLM with the sampling temperature $T$; 
next, to arrive at an effective temperature $T'$, it performs post-hoc scaling by a factor $\alpha = \frac{T}{T'} - 1$.
In principle, temperature scaling may not be necessary if $\hat{p}(y \,|\, x)$ accurately estimated the true probability $p(\cdot \,|\, x)$. 
However, in practice due to imperfect approximation and finite sample size errors, we find it useful to employ.
Indeed, temperature scaling has also been found to be beneficial in prior MBR works~\citep{yan2022dc}.

\subsection{Extension to multi-partite ranking}
\label{s:ranking}
Our metric-aware decoding proposal also applies to scoring tasks, where the label space $Y$ is discrete, e.g. $\{1, \ldots, K\}$, but we require the LLM to predict real-valued scores $\hat{y}(x) \in \mathbb{R}$ for each prompt $x$ such that prompts with higher labels receive a higher score. 
One typically measures the performance of $\hat{y}(x)$ using a pairwise ranking metric such as 
the multi-partite area under the ROC curve (AUC-ROC)~\citep{Uematsu2015}:
\begin{align}
    \text{AUC-ROC}(\hat{y}) = 1 - \mathbb{E}\Big[c_{y,y'}\cdot \mathds{1}(\hat{y}(x) < \hat{y}(x'))\,\Big|\, y > y' \Big],
    \label{eq:auc}
\end{align}
which penalizes the scorer $\hat{y}$ by  $c_{y,y'}$ whenever it mis-ranks a pair $(x,  x')$ with $y > y'$.
In experiments, we refer to AUC-ROC as AUC for brevity.

Despite AUC-ROC being non-decomposable (not a summation of per-example results), 
\citet{Uematsu2015}[Corollary 1] show that when the costs are the difference between the labels, i.e., $c_{y, y'} = |y - y'|$, 
the optimal scorer admits a closed-form solution given by the expected label under distribution $p(\cdot|x)$:~ $\hat{y}^*(x) = \mathbb{E}_{y \sim p(\cdot|x)}\left[ y \right]$. One can thus readily apply our RAIL approach to estimate this solution from sampled responses.
Moreover, from the Neyman-Pearson lemma, the same optimal solution applies to the AUC-PR evaluation metric \cite{clemenccon2009nonparametric}.

\vspace{-5pt}
\section{Experiments and Discussion}
\label{s:exp}
\vspace{-5pt}
We experimentally evaluate our proposed approach %
on  NLP tasks with different evaluation metrics.%

\noindent\textbf{Datasets.}\ 
We use two datasets: (i) Semantic Textual Similarity Benchmark (\emph{STSB}) \citep{Cer_2017}, which comprises of sentence pairs human-annotated with a similarity score from 0 to 5; %
since this is a regression task, we evaluate with the root mean squared error. (ii)   \emph{US Amazon reviews}, %
where we aim to predict the 5-star rating for a product review \citep{ni-etal-2019-justifying};
since the task is in the form of ordinal regression, we use mean absolute error as the  evaluation metric \citep{NIPS2017_c86a7ee3}.
We list the prompts used in Table~\ref{table:prompts} (Appendix). In each case, we evaluate on samples of $1500$ examples.%

\begin{table}
\footnotesize
\begin{center}
\tabcolsep=0.1cm
\begin{tabular}{l l @{\hskip 0.1in} c @{\hskip 0.15in} ccc @{\hskip 0.15in} ccc @{\hskip 0.15in}  cc}
\toprule 
& \multirow{2}{*}{\shortstack[l]{model\\size}} & \multirow{2}{*}{\shortstack[l]{greedy\\decode}} & & %
RAIL\\
 & & & argmax & mean\\
\midrule
\multirow{3}{*}{\shortstack[c]{STSB\\(RMSE$\downarrow$)}} & XXS & 1.078 & 1.448 & \textbf{1.028}\\
 & S &  0.685 & 1.019 & \textbf{0.649}\\
 & L & 0.628 & 0.989 & \textbf{0.610}\\
\midrule
 & & & argmax & mean\\
\midrule
\multirow{3}{*}{\shortstack[c]{STSB\\(AUC$\uparrow$)}} & XXS & 0.797 & 0.632 & \textbf{0.889}\\
 & S &  0.895 & 0.820 & \textbf{0.953}\\
 & L & 0.905 & 0.827 & \textbf{0.961}\\
\midrule
 & & & argmax & median \\
\midrule
\multirow{3}{*}{\shortstack[l]{Amazon\\reviews\\(MAE$\downarrow$)}} & XXS & 0.495 & 0.826		& 	\textbf{0.474}\\
 & S & 0.301 & 0.444	& 		\textbf{0.285}\\
 & L & 0.294 & 0.541	& 		\textbf{0.291}\\
\bottomrule 
\end{tabular}
\end{center}
\caption{Comparison of inference strategies %
on PaLM-2 models for different datasets and metrics. We draw $16$ samples  %
with an effective temperature of $T=\frac{1}{4}$ (via post-hoc scaling). In Appendix \ref{app:additional-expts}, we report results for variants of MALI with no post-hoc scaling (Table~\ref{tab:results-all-temps}), and results of tuning the temperature using a held-out set, along with confidence intervals (Table~\ref{tab:results-tune-temp}).}
\label{tab:results-all}
\end{table}

\noindent\textbf{Models.}\
We consider two instruction-tuned model families: PaLM-2~\citep{palm2} and FLAN-T5~\citep{chung2022scaling}.
We report results across different model sizes and temperatures.
Unless otherwise stated, we fix the number of samples to $K = 16$, %
and the top-$k$ parameter in decoding to $40$~\citep{fan2018hierarchical}.~\\[-10pt]

\noindent\textbf{Methods.}\ We evaluate the following methods: (i) greedy decoding, (ii) a baseline inspired from the self-consistency decoding of sampling $K$ candidates  and picking the one with the maximum likelihood (argmax) \citep{wang2023selfconsistency}, (iii) the proposed RAIL approach  on the same $K$ samples, and (iv) RAIL with temperature scaling (\S\ref{ssec:post-hoc-temp}). 
For (iv), we choose $\alpha$ yielding effective temperature $\frac{1}{4}$.

\noindent\textbf{Metric-aware inference helps.}\ 
In Table~\ref{tab:results-all}, we report results across datasets and model sizes. 
We notice that RAIL consistently improves over baselines. %
To better measure the sensitivity of the results to the choice of temperature, we report additional results in Table~\ref{tab:results-tune-temp} in Appendix \ref{app:additional-expts}, where we use a held-out validation set to tune the temperature, and find the trends to be consistent. %

\noindent\textbf{Sampling versus enumeration.} 
So far, when estimating the maximizer to equation~\eqref{eq:expected-utility}, we have used sampling from the LM distribution (see \S\ref{s:approx_bayes_opt}). 
Alternatively, if the targets are from a narrow interval (e.g., on STSB, the values are in the interval $[0, 5]$), one can score the model for targets enumerated at fixed intervals (e.g. $0, 0.5, 1.0, \ldots, 5.0$), and compute estimates for solutions in Table~\ref{tbl:decision_rules}. %
In Table~\ref{tbl:STSB_FLANT5}, we report results from FLAN-T5 on the STSB dataset for RAIL with both sampling and enumeration based estimates, where the latter is based on $11$ equally spaced targets.
Both sampling and enumeration lead to RAIL improving over choosing the most likely target, with sampling having an edge. The reason sampling performs better than enumeration may be that sampling is able to better explore the high density regions of the output probability space, as we detail in Appendix~\ref{a:why_sampling_outperforms}.

\begin{table}
\small
    \centering
    \renewcommand{\arraystretch}{1.25}

    \begin{tabular}{@{}lccc@{}}
        \toprule
         \textbf{model} & \textbf{greedy} & \textbf{enumeration} & \textbf{sampling}\\ %
\toprule
\midrule
FLAN-T5 S & 2.102 & 1.551 & \textbf{1.508}\\
FLAN-T5 L & 0.675 & 0.640 & \textbf{0.611}\\
FLAN-T5 XL & 0.713 & 0.741 & \textbf{0.676}\\
  \bottomrule\\[-7pt]
\end{tabular}
    \caption{RMSE on STSB with FLAN-T5 across RAIL variants (enumeration vs sampling). The sampling approach uses a temperature of $0.5$.} %
    \vspace{-15pt}
    \label{tbl:STSB_FLANT5}
\end{table}

\noindent\textbf{Role of model size.}\ 
We find that the benefit from our technique reduces as the models increase in size. %
This sometimes coincides with a lowering entropy in predictions with increasing model size (see, e.g., results on Amazon in  Table~\ref{tbl:entropy} in Appendix).  %
We note this is consistent with prior works on MBR, which observed that as the model gets better, the optimal decision rule for EM (approximated by greedy decoding) performs comparable to the that for other metrics \citep{Schluter2012}.
We stress that the gains we get with  small and medium-sized models are still of large practical importance, 
especially in applications where deploying very large models is prohibitively expensive.

\section{Conclusions}
In this work we show how adopting MBR to regression and scoring tasks, and thus utilizing the output distribution modeled by LLMs in the form of our RAIL methods can bring improvements. 
Our work also points at importance of further understanding and improving of the quality of model output distribution and calibration, which we found crucial for the RAIL methods to work well.
In the future, we wish to extend our approach to other less-explored evaluation metrics in the MBR literature.

\section{Limitations}
There are multiple limitations of our work.
First, we evaluate our proposed methods on multiple text datasets with numerical and text targets, however, many more types of outputs can be considered, including the time series targets.
Next, it would be interesting to more systematically analyze how to efficiently solve the objective from \eqref{eq:bayes-opt-empirical-weighteed} over many samples for text outputs for metrics like $F_1$ or BLEU, e.g. by means of dynamic programming.
We also note that the datasets considered in this work are restricted to English. It would be interesting to expand the explorations to datasets in other languages.

\section{Ethics Statement}
All datasets used in this work are publicly available.
No additional user data was collected or released as part of this work.
All models used are publicly available and already pretrained, and no fine-tuning was conducted for any experiments.
Instead, all experiments relied on running inference experiments with the models over several thousands of examples.
Thus, the CO-2 footprint of this paper is minimal.
We do not foresee any significant risks associated with this paper other than improving performance on tasks which are harmful.

\section{Acknowledgements}
We are thankful to Changsheng Jiang for experiments and analyses of early versions of our technique.
We are also thankful to Ziwei Ji and Shankar Kumar for helpful feedback and comments.

\bibliography{main}
\ifarxiv
    \bibliographystyle{plainnat}
\fi

\newpage

\appendix

\section{Further related work}
\label{s:relatedwork}
\paragraph{Minimum Bayes risk decoding.}\  As noted in the introduction, prior work on MBR has considered optimizing for common metrics in the machine translation and text generation literature. The closest to our paper is the work of
\citet{wang2023selfconsistency}, who considered sampling from the model distribution using chain-of-thought (CoT) prompting, and showed how majority vote improves over the baseline on arithmetic and reasoning tasks.

Other works explored different aspects of MBR, including: the role of the sampling algorithms \citep{freitag-etal-2023-epsilon,cheng2023faster}, the interaction with label smoothing \citep{yan2022dc}, and how it generalizes other techniques \citep{suzgun2022follow,bertsch-etal-2023-mbr}.
\citet{finkelstein2024mbr} recently considered distillation of MBR solution to a student model, so as to avoid the overhead induced by MBR at inference time. 

A recent work also applied LLMs to time series forecasting, and constructed the final predictions by computing quantiles (e.g., median) over the samples \citep{gruver2023large}. One of the evaluation metrics for time series forecasting is the mean absolute error, for which the median can be shown to be a Bayes optimal decision rule (see Table \ref{tbl:decision_rules}).

\paragraph{Fine-tuning for target task alignment.}\ Previous works have considered approaches for aligning the models for target datasets. 
This includes fine-tuning of soft prompts on target datasets without losing generalization to other tasks \citep{wang2023twostage}, and general fine-tuning on carefully tailored datasets for improved model robustness \citep{li-etal-2023-large}.
In our work, we focus on zero-shot setting where no fine-tuning is conducted.

\paragraph{Fine-tuning for numerical tasks.}\ Autoregressive fine-tuning of LLMs on numerical tasks with CoT has been found effective \citep{liu2023goat}. 
One line of work for modeling predictive tasks with pre-trained Transformer-based models is to add a regression head on top of the transformed/pooled encoded input tokens and fine-tune the resulting model on numerical targets using a regression loss. 
This is an approach which has been employed for encoder-based models (e.g., BERT), and has also been applied to encoder-decoder (e.g., T5) models \citep{liu2022enct5}, and these approaches could be extended to decoder models too. 
In a similar work, an embedding was extracted from a decoder model fine-tuned on modified attention mask and additional tasks \citep{behnamghader2024llm2vec}.
In this paper, we focus on the zero shot approaches, and  we leave training approaches for future work.

\begin{table*}
\footnotesize
\begin{center}
\scalebox{0.85}{
\tabcolsep=0.1cm
\begin{tabular}{l l @{\hskip 0.15in} c @{\hskip 0.2in} ccc @{\hskip 0.2in} ccc @{\hskip 0.2in}  ccc}
\toprule 
& \multirow{2}{*}{\shortstack[l]{model\\size}} & \multirow{2}{*}{\shortstack[l]{greedy\\decode}}& \multicolumn{3}{c}{T=$0.25$}  & \multicolumn{3}{c}{T=$0.5$}  & \multicolumn{3}{c}{T=$1.0$}\\
 & & & argmax & mean & w-mean & argmax &  mean & w-mean & argmax & mean & w-mean \\
\midrule
\multirow{3}{*}{STSB} & XXS & 1.078 & 1.126 & 1.043 & 1.028 & 1.241 & 1.021 & 0.992 & 1.448 & 1.007 & \textbf{0.978}\\
 & S &  0.685 & 0.787 & 0.643 & 0.649 & 0.908 & \textbf{0.636} & 0.642 & 1.019 & 0.641 & 0.641\\
 & L & 0.628 & 0.729 & 0.592 & 0.610 & 0.852 & 0.582 & 0.586 & 0.989 & \textbf{0.580} & \textbf{0.580}\\
\midrule
&  & \multirow{2}{*}{\shortstack[l]{}}& \multicolumn{3}{c}{T=$0.25$}   & \multicolumn{3}{c}{T=$0.5$}  & \multicolumn{3}{c}{T=$1.0$}\\
 & & & argmax & median & w-median &  argmax &  median & w-median & argmax & median & w-median \\
\midrule
\multirow{3}{*}{\shortstack[l]{Amazon\\reviews}} & XXS & 0.495 & 0.509		& 	0.484& 	\textbf{0.474}& 	0.624		& 	0.485	& 0.487& 	0.826	& 		0.493& 	0.493\\
 & S & 0.301 & {0.290}	& 		0.297	& \textbf{0.285}& 	0.329		& 	0.300	& 0.297	& 0.444		& 	0.299& 	0.299\\
 & L & 0.294 & 0.318	& 		0.293 & 	\textbf{0.291}& 	0.380		& 	0.294	& {0.293}& 	0.541& 			0.298& 	0.295\\
\midrule
&  & \multirow{2}{*}{\shortstack[l]{}} & \multicolumn{3}{c}{T=$0.25$}  & \multicolumn{3}{c}{T=$0.5$}  & \multicolumn{3}{c}{T=$1.0$}\\
 & & & argmax & $F_1$ & w-$F_1$ & argmax & $F_1$ & w-$F_1$ & argmax & $F_1$ & w-$F_1$ \\
\midrule
\multirow{3}{*}{Trivia-QA} & XXS & 0.314 & 
0.300 &	0.319&	0.318&
 0.255&0.323&	\textbf{0.326}&
 0.178	&0.307&	0.304  \\
 & S & 0.620 & 
0.656&0.626	&\textbf{0.678} & 
 0.658&	0.641&	0.662&	
 0.636	&	0.650 & 0.650\\
 & L & 0.886 &
\textbf{0.888}	&0.886	&\textbf{0.888}&
 \textbf{0.888}&	0.883	&0.887	&
 0.887	&	0.880&	0.885\\
\bottomrule 
\end{tabular}
}
\end{center}
\caption{Root mean squared error (RMSE) on STSB dataset (the lower the better), Mean absolute error (MAE) on Amazon reviews dataset (the lower the better), and $F_1$ metrics on Trivia-QA dataset (the higher the better) from PaLM-2 models of varying size. We report different methods of inference across different temperatures. For the weighted approaches, we fix the sampling temperature to $T=1$ and accordingly vary the $\alpha$ in \eqref{eq:bayes-opt-empirical-weighteed} so as to arrive at the effective temperature equal to the value reported.}
\label{tab:results-all-temps}
\end{table*}
\begin{table}
\small
    \centering
    \renewcommand{\arraystretch}{1.25}
    \begin{tabular}{@{}lcc@{}}
        \toprule
         \textbf{model} & w/ pairs & w/o pairs\\
\toprule
\midrule
PaLM-2 XXS & $0.302$ & $0.295$\\
PaLM-2 XS &  $0.678$ & $0.670$\\
PaLM-2 L & $0.886$ & $0.887$\\
  \bottomrule\\[-7pt]
\end{tabular}
    \caption{Performance of RAIL (as evaluated by $F_1$) on TriviaQA with and without the inclusion of concatenated pairs in the candidate set.}
    \label{tbl:candidate_set}
\end{table}
\section{Additional results on $F_1$ maximization on Trivia QA}
\label{app:triviaqa}
We extend our approach to the $F_1$
 score evaluation metric. Consider a reading comprehension task, where the $F_1$ score is the evaluation metric $m(y, \hat{y})$, defined by the harmonic mean of $\text{recall}(y, \hat{y}) = \frac{|y \cap \hat{y}|}{|y|}$ and $\text{precision}(y, \hat{y}) = \frac{|y \cap \hat{y}|}{|\hat{y}|}$. To illustrate the task, suppose for the question {\tt ``What is the hottest month in the year''}, the responses and associated probability from an LM are \{{\tt``July''}: 0.25,\, {\tt``July 2023''}: 0.23,\, {\tt``Month of July''}: 0.24,\, {\tt``May''}: 0.28\}. The mode of this distribution is {\tt ``May''}; whereas the maximizer of \eqref{eq:expected-utility} is {\tt ``July''}.

To optimize the $F_1$ metric, we solve  \eqref{eq:bayes-opt-empirical-C} over a candidate set $C$, which we choose to contain the $K$ samples and additional targets derived from them. %
\begin{align}
    \hat{y}(x) \in \argmax_{y' \in C}\, \sum_{i=1}^K m(y_i, y').
    \label{eq:bayes-opt-empirical-C}
\end{align}
While the $F_1$ score does not admit a closed-form solution, as is the case for the metrics listed in Table~\ref{tbl:decision_rules}, we make an observation that its formulation allows for introducing a different form of efficiency.
In particular, we notice that due to the trade-off between precision and recall in the $F_1$ score formulation, the following candidate set construction can lead to increasing recall at the expense of precision, thus providing a way to cheaply enumerate additional reasonable candidates.
\begin{table*}
  \centering
    \scalebox{0.73}{
    \begin{tabular}{p{3cm}p{15cm}}
    \toprule
    {\textbf{Dataset}} & {\textbf{Prompt}} \\
    \midrule
    STSB & What is the sentence similarity between the following two sentences measured on a scale of 0 to 5: \{Sentence \#1\}, \{Sentence \#2\}. The similarity measured on a scale of 0 to 5 with 0 being unrelated and 5 being related is equal to \\ 
    \midrule
    Amazon reviews & What is the rating corresponding to the following review in the scale of 1 to 5, where 1 means negative, and 5 means positive? Only give a number from 1 to 5 with no text. Review: \{Review\} Rating:\\
    \bottomrule
    \end{tabular}
    }
    \caption{Prompts used for different datasets. Curly braces denote inputs specific to an input example.}
    \label{table:prompts}
\end{table*}

\noindent\textbf{Candidate set construction.}\ 
One simple choice for the candidate set $C$ could be take the $K$ sampled outputs, i.e., $C = \{y_1, \ldots, y_K\}$. One may additionally include in this set transformations on each $y_i$ or new candidates formed from combining two or more of the samples.

For reading comprehension or question-answering applications, where the output is a list
of keywords that constitute an answer to a question, one may additionally include samples formed by concatenating pairs of sampled outputs, i.e., ${\tt concat}(y_i, {\tt delim}, y_j), \forall i \ne j$. These concatenated answers have the effect of increasing recall, at the cost of lower precision.
We follow that procedure for the Trivia-QA experiments. %

In Table \ref{tab:results-all-temps}, we provide results on Trivia-QA reading comprehension task \citep{triviaqa} with the proposed $F_1$-aware inference strategy.

To additionally analyze the effectiveness of the candidate set augmentation, in Table~\ref{tbl:candidate_set} we compare the performance of RAIL (specifically the temperature scaled variant) with and without the inclusion of concatenated pairs in the candidate set. 
For both the XXS and S models, the inclusion of concatenated pairs is seen to yield a significant improvement in $F_1$-score.

\section{Additional details}
\label{app:additional}
In Table~\ref{table:prompts} we report the prompts we used in our experiments for zero-shot inference.

For all datasets, we use validation splits, and where not available, we use the first $1500$ examples from the train split. 

The datasets are publicly available, for example from the \url{tensorflow.org} platform:
\begin{itemize}
    \item \url{https://www.tensorflow.org/datasets/catalog/glue\#gluestsb},
    \item \url{https://www.tensorflow.org/datasets/catalog/amazon_us_reviews},
    \item \url{https://www.tensorflow.org/datasets/catalog/trivia_qa}.
\end{itemize}

\section{Additional experiments}
\label{app:additional-expts}
In Table~\ref{tbl:entropy} we report empirical entropy estimates as measured based on the $16$ samples generated from the model. 
We find that entropy decreases as model size increases. We observe a particularly sharp decrease in entropy for the Amazon reviews and Trivia-QA datasets, where for larger model sizes we don't find improvements from RAIL approaches.

In Table~\ref{tab:results-all-temps} we report RMSE on STSB dataset, MAE on Amazon reviews dataset, and $F_1$ metrics on Trivia-QA dataset from PaLM-2 models of varying size across multiple temperature values. We find improvements over baselines on STSB and Amazon reviews datasets for most temperatures. For Trivia-QA, we find improvements for XXS and S models for some temperatures, and for L, we don't find a difference from our methods due to low entropy in the responses (see Table~\ref{tbl:entropy}).
In Table~\ref{tbl:pearson} we additionally report Pearson correlation metrics on STSB, confirming the results of RAIL improving over autoregressive inference.
Lastly, in Table~\ref{tbl:auc} we report cost weighted multi-class AUC-ROC with costs corresponding to the difference between the annotated labels: $|y_1-y_2|$.
We find on both STSB and Amazon reviews datasets that the optimal decision rule (mean over the distribution) improves over the baselines.

\begin{table}
\small
    \centering
    \renewcommand{\arraystretch}{1.25}

    \begin{tabular}{@{}lccc@{}}
        \toprule
         \textbf{model} & STSB & Amazon & Trivia-QA\\
\toprule
\midrule
PaLM-2 XXS & $1.141$ & $1.064$ & $1.328$\\
PaLM-2 XS &  $1.055$ & $0.753$ & $0.475$\\
PaLM-2 L & $0.976$ & $0.361$ & $0.186$\\
  \bottomrule
\end{tabular}
    \caption{Empirical entropy across model sizes and datasets.}
    \label{tbl:entropy}
\end{table}

\begin{table}
\small
\begin{center}
\tabcolsep=0.08cm
\begin{tabular}{r @{\hskip 0.3in} c @{\hskip 0.3in} c @{\hskip 0.3in}  c}
\toprule 
samples &  XXS  & S  & L\\
\midrule
(Greedy Decode) & 1.078 & 0.685 & 0.628\\
\midrule
2 & 1.044 & 0.679 & 0.624\\
4 & 1.036 & 0.669 & 0.613\\
6 & 1.031 & 0.664 & 0.607\\
8 & 1.028 & 0.660 & 0.603\\
10 & 1.025 & 0.657 & 0.601\\
12 & 1.024 & 0.655 & 0.600\\
14 & 1.022 & 0.653 & 0.599\\
16 & 1.021 & 0.652 & 0.598\\
\bottomrule 
\end{tabular}
\end{center}
\caption{RMSE as a function of the number of samples on STSB across PaLM-2 models of varying size. Results for temperature $T=0.25$.}
\label{tbl:samples}
\end{table}
\begin{table*}
\small
    \centering
    \renewcommand{\arraystretch}{1.25}

    \begin{tabular}{@{}llc @{\hskip 0.2in} cc @{\hskip 0.2in}  cc@{\hskip 0.2in}  cc@{}}
        \toprule
         &  
         \multirow{2}{*}{\shortstack[l]{model\\size}}&   \multirow{2}{*}{\shortstack[l]{greedy\\decode}}  &  \multicolumn{2}{c}{T=$0.25$}  & \multicolumn{2}{c}{T=$0.5$}  & \multicolumn{2}{c}{T=$1.0$}\\
         & & & argmax & mean & argmax & mean & argmax & mean\\
\toprule
\midrule
\multirow{3}{*}{\shortstack[l]{STSB}} 
& XXS & $0.797$ & $0.755$ & $0.882$ & $0.714$ & $0.890$ & $0.632$ & $0.889$\\
& XS & $0.895$ & $0.870$ & $0.950$ & $0.843$ & $0.954$ & $0.820$ & $0.953$\\
& L & $0.905$ & $0.885$ & $0.948$ & $0.859$ & $0.959$ & $0.827$ & $0.961$\\
\midrule
\multirow{3}{*}{\shortstack[l]{Amazon\\reviews}} & XXS & $0.87$ & $0.894$ & $0.925$ & $0.866$ & $0.94$ & $0.788$ & $0.942$\\
& XS &  $0.9$ & $0.91$ & $0.925$ & $0.914$ & $0.941$ & $0.9$ & $0.958$\\
& L & $0.925$ & $0.922$ & $0.951$ & $0.906$ & $0.962$ & $0.837$ & $0.964$\\
  \bottomrule
\end{tabular}
    \caption{Cost-weighted multi-partite AUC metrics on STSB and Amazon datasets (\emph{the higher the better}). RAIL methods improve over the baselines. See \S\ref{s:ranking} for the definition of AUC we use. We assume costs to correspond to the difference between the annotated labels: $|y_1-y_2|$.}
    \label{tbl:auc}
\end{table*}

\begin{table*}
\small
    \centering
    \renewcommand{\arraystretch}{1.25}

    \begin{tabular}{@{}lc @{\hskip 0.2in} cc @{\hskip 0.2in}  cc@{\hskip 0.2in}  cc@{}}
        \toprule
         \textbf{model}&   \multirow{2}{*}{\shortstack[l]{greedy\\decode}}  &  \multicolumn{2}{c}{T=$0.25$}  & \multicolumn{2}{c}{T=$0.5$}  & \multicolumn{2}{c}{T=$1.0$}\\
         & & argmax & mean & argmax & mean & argmax & mean\\
\toprule
\midrule
PaLM-2 XXS & $0.767$	& 0.738 & $\mathbf{0.790}$	& 0.670 & $\mathbf{0.790}$	& 0.544	& $0.786$\\
PaLM-2 XS &  $0.898$	& 0.878 & $\mathbf{0.915}$	& 0.852 & $0.913$		& 0.821 & $0.910$\\
PaLM-2 L & $0.909$	& 0.893 & $0.920$	& 0.881 & $0.922$		& 0.860 & $\mathbf{0.923}$\\
  \bottomrule
\end{tabular}
    \caption{Pearson correlation metrics on STSB. RAIL methods improve over the baselines.}
    \label{tbl:pearson}
\end{table*}

In Table~\ref{tbl:samples}, we report the impact of the number of samples on the results.
We note that there is an improvement in the results with the increase in the number of samples, however beyond 8 samples there is a diminishing improvement in practice.
On STSB with temperature $\frac{1}{4}$, even with as few as \emph{two} samples, our method starts to show improvements over greedy decoding.

In Table \ref{tab:results-tune-temp}, we report results for PaLM-2 models for RMSE on STSB when tuning the temperature parameter using a held-out set. 

\section{Why does sampling outperform enumeration?}
\label{a:why_sampling_outperforms}
In this section we explicate why sampling can outperform enumeration.
For easier reference, we first summarize what sampling and enumeration specifically mean:
\begin{itemize}
\item with the \emph{sampling} strategy, we evaluate the average metric in \eqref{eq:bayes-opt-empirical} using $K$ samples drawn from the predictive distribution through temperature sampling. 
\item with the \emph{enumeration} strategy, we score $K$ fixed targets from a uniform grid $G$, and replace the average metric in equation~\eqref{eq:bayes-opt-empirical} with the estimate $$\sum_{g \in G} p(g) \cdot m(y, g) / \sum_{g \in G} p(g).$$
\end{itemize}
Now, a possible reason sampling performs better than enumeration can be that sampling is able to better explore the high density regions of the output probability space. For example, if the predictive distribution is concentrated in a tiny region of the output space, with the sampling strategy, most of the samples we use to estimate the optimal solution will be from this region. In contrast, with the enumeration strategy, most of the enumerated outputs will be from outside this region, and may prove not useful for estimating the optimal solution.

For illustrative purposes, consider an extreme example for the STSB regression setup (where the output is a real number in $[0, 5]$). Suppose the predictive distribution is a mixed probability distribution whose density is concentrated in a narrow region centered at $0.7$, and is near-uniform on all other targets. Since our enumeration strategy only considers the grid points $G = \{0.0, 0.5, 1.0, \ldots, 5.0\}$, due to uniform probabilities over all these values it outputs: 
$$\sum_{g \in G} p(g) \cdot g / \sum_{g \in G} p(g) = 2.5.$$
With the sampling approach, all K samples will be drawn with high probability from the vicinity of $0.7$, and so, its output is:
$$\frac{1}{K} \sum_k \hat{y}_k \approx  0.7.$$

We would also like to note that both scoring and sampling improve over baselines, showing that both alternatives can make good use of the $\hat{p}(.|x)$. We also note that p may not be perfectly approximated by $\hat{p}$ due to various reasons, including the optimization, capacity, limited fine-tuning data and objectives used (e.g. label smoothing used in the pre-training objective).

\section{Computational complexity of sampling-based RAIL}

Note that sampling can be done efficiently by caching the Transformer activations for the input prefix when generating different targets.
In practice, when the prefix is long compared to the generated targets, a forward pass for the prefix tends to take most of the compute time. 
Note that is the case for scoring and regression tasks (the focus of our work), since the target score can be just a few tokens length, whereas the prefix can be long as it contains the input text.

Moreover, we generate multiple samples simultaneously, and so, we do not incur a higher cost from generating multiple targets.

\begin{table*}
\footnotesize
\begin{center}
\tabcolsep=0.1cm
\begin{tabular}{ l @{\hskip 0.1in} c @{\hskip 0.15in} ccc @{\hskip 0.15in} ccc @{\hskip 0.15in}  cc}
\toprule 
\multirow{2}{*}{\shortstack[l]{model\\size}} & \multirow{2}{*}{\shortstack[l]{greedy\\decode}} & & %
RAIL\\
 & & argmax & mean\\
\midrule
XXS &  1.047$\pm$0.004 & 1.447$\pm$0.007 & \textbf{0.967$\pm$0.004} \\
 S &  0.683$\pm$0.002 & 1.017$\pm$0.005 & \textbf{0.639$\pm$0.003}\\
 L & 0.628$\pm$0.003 & 0.988$\pm$0.004 & \textbf{0.578$\pm$0.002}\\
\bottomrule 
\end{tabular}
\end{center}
\caption{Comparison of inference strategies %
on PaLM-2 models for RMSE on STSB when tuning the temperature on a held-out set. We draw $16$ samples.  %
We use $\frac{1}{3}$ of the evaluation set for selecting the temperature from $\{0.25, 0.5, 0.75, 1, 2.5, 5, 7.5\}$, and use the remaining $\frac{2}{3}$ of the evaluation set for evaluation. We draw 10 random splits to obtain 95\% confidence intervals. We confirm that the improvements that RAIL offers over baselines are indeed significant, and that when tuning the temperatures on the held-out set, the improvements hold.
}
\label{tab:results-tune-temp}
\end{table*}

\end{document}